\documentclass{article}

\usepackage{PRIMEarxiv}
\usepackage[utf8]{inputenc} 
\usepackage[T1]{fontenc}    
\usepackage{hyperref}       
\usepackage{url}            
\usepackage{booktabs}       
\usepackage{amsfonts}       
\usepackage{nicefrac}       
\usepackage{microtype}      
\usepackage{lipsum}
\usepackage{fancyhdr}       
\usepackage{graphicx}       
\usepackage{amsmath,amssymb}
\usepackage{float}
\usepackage{wrapfig}        
\usepackage{subcaption}
\usepackage{svg}
\graphicspath{{media/}}     

\numberwithin{equation}{section}
\numberwithin{figure}{section}

\title{Prediction of the evolution of the nuclear reactor core parameters using artificial neural network

}
\author{
  Krzysztof Palmi \\
  Faculty of Physics \\
  Warsaw University of Technology \\
  \texttt{krzysztof.palmi.stud@pw.edu.pl} \\
  \And
  Wojciech Kubi{\'n}ski \\
  National Centre  \\
  for Nuclear Research \\
  \texttt{wojciech.kubinski@ncbj.gov.pl} \\
  \And
  Piotr Darnowski \\
  Institute of Heat Engineering \\
  Warsaw University of Technology \\ 
  \texttt{piotr.darnowski@pw.edu.pl} \\
}

\begin{document}
\maketitle
\begin{abstract}

The aim of the research was to design, implement and investigate an Artificial Intelligence (AI) algorithm, namely an Artificial Neural Network (ANN) that can predict the behaviour of selected parameters of a nuclear reactor core.

A nuclear reactor based on MIT BEAVRS benchmark was used as a typical power generating Pressurized Water Reactor (PWR). The PARCS v3.2 nodal-diffusion core simulator was used as a full-core reactor physics solver to emulate the operation of a reactor and to generate training, and validation data for the ANN. 

The ANN was implemented with dedicated Python 3.8 code with Google's TensorFlow 2.0 library. The effort was based to a large extent on the process of appropriate automatic transformation of data generated by PARCS simulator, which was later used in the process of the ANN development.

Various methods that allow obtaining better accuracy of the ANN predicted results were studied, such as trying different ANN architectures to find the optimal number of neurons in the hidden layers of the network. Results were later compared with the architectures proposed in the literature. For the selected best architecture predictions were made for different core parameters and their dependence on core loading patterns. 

In this study, a special focus was put on the prediction of the fuel cycle length for a given core loading pattern, as it can be considered one of the targets for plant economic operation. For instance, the length of a single fuel cycle depending on the initial core loading pattern was predicted with very good accuracy (>99\%). 

This work contributes to the exploration of the usefulness of neural networks in solving nuclear reactor design problems. Thanks to the application of ANN, designers can avoid using an excessive amount of core simulator runs and more rapidly explore the space of possible solutions before performing more detailed design considerations.

\end{abstract}

\keywords{artificial neural network \and nuclear reactor \and batch learning \and TensorFlow \and PARCS \and BEAVRS}



\section{Introduction}
\label{sec:intro}


\subsection{Artificial Intelligence in nuclear engineering}

Nuclear power is an increasingly significant source of electricity in the World. In 2021, nuclear reactors accounted for about 10\% of globally produced electricity and were the second-largest source of low-emission electricity~\cite{IEA}. With the development of this energy sector, the demand for optimizing processes related to the design and management of nuclear reactors is increasing. Moreover, increasing easily accessible computational resources allow engineers to propose more and more sophisticated methods for process optimization.

Over the past decades Artificial Intelligence (AI) gained a lot of interest in nuclear engineering due to the possible modernization of software in existing and new nuclear reactor technologies \cite{arndt2015}. The need of addressing the new solutions was stated by the International Atomic Energy Agency (IAEA) in 1986~\cite{marchesi1986nuclear} and as a result, a wide variety of methods were already developed in this field, some of which focused mostly on technical aspects such as reducing radiation exposure to personnel, enhancing the reliability of equipment and others focusing mostly on economical aspects such as optimization of the maintenance schedule or improving plant availability. When it comes to different algorithms used for such problems it is needed to mention naive Bayes classifiers in bayesian-based isotope identification~\cite{SULLIVAN2015298}, optimization methods like Genetic Algorithms  \cite{KUBINSKI2021108153}\cite{KubinskiDarnowskiChec+2021+147+151}, and most common machine learning (ML) algorithms such as decision trees, support vector machines (SVM) and its use in classification of uranium waste~\cite{HATA2015143}, and deep learning models such as artificial neural network (ANN) and convolutional neural network (CNN) - both having a contribution to nuclear engineering field with identification of accident scenarios in nuclear power plants using ANNs~\cite{SANTOSH2009759} or approximation of flows in a reactor using CNNs~\cite{10.1145/2939672.2939738} (more detailed state of art can be found in~\cite{GOMEZFERNANDEZ2020110479}). 

One of the solutions to the reactor design process optimization is the use of ANNs, on which this research is focused and which could calculate different parameters of the nuclear reactor core cycle in just seconds compared to running a full advanced simulation.

\subsection{Artificial neural networks}  

Neural networks, often known as Artificial Neural Networks (ANNs), are computational models based on the biological neural network found in the brain. They have been designed to create a complex solution based on artificial intelligence that allows for nonlinear regression and classification problems.

\begin{figure}[htp]
  \centering
  \includegraphics[width=0.8\textwidth]{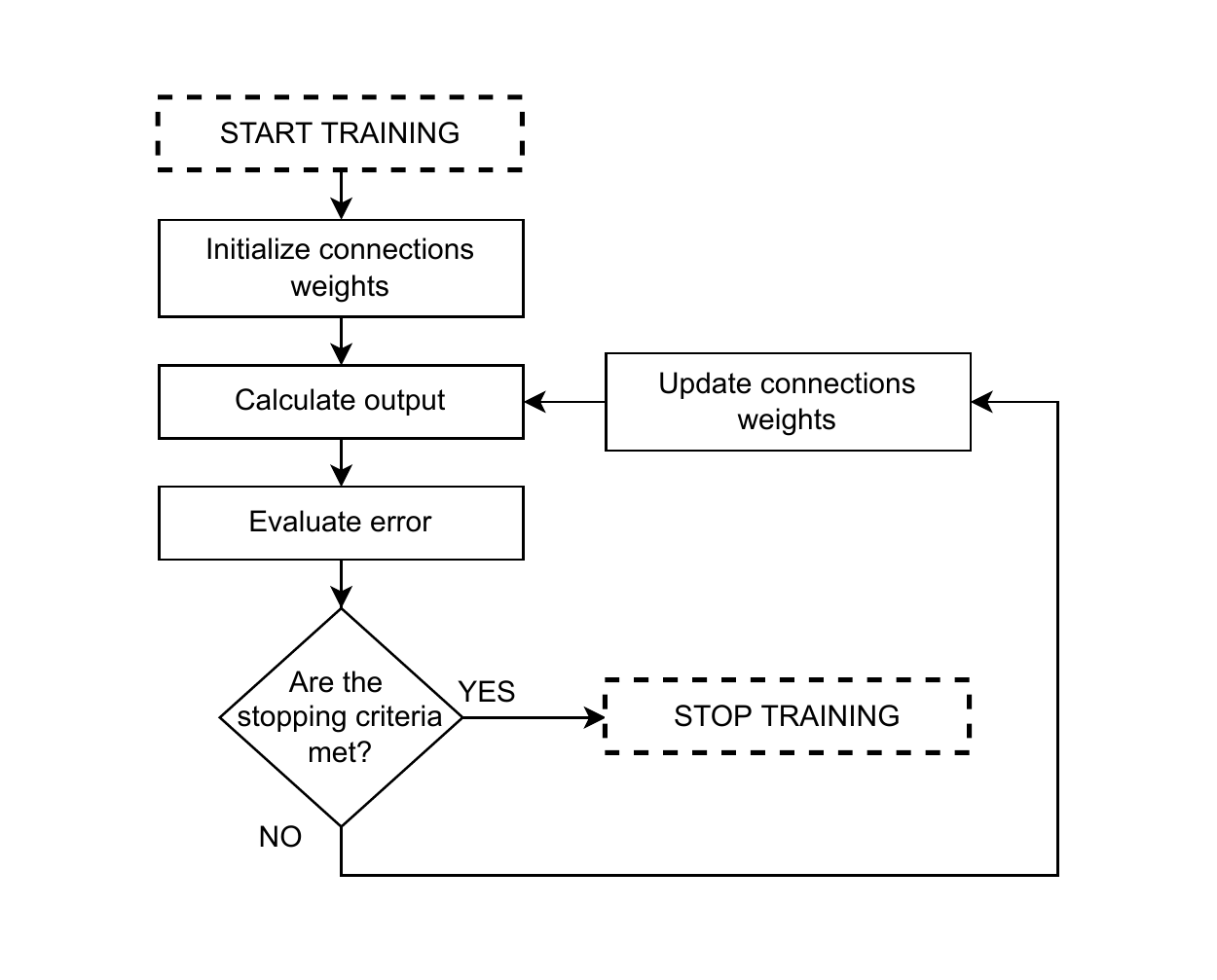}
  \caption{Backpropagation algorithm described in \cite{kelley1960gradient}.}
  \label{fig:backpropagation_diagram}
\end{figure}

Such a network consists of units called neurons, which are connected like synapses that transmit information to other biological neurons. In the case of a simple type of Multilayer Perceptron (MLP) neural network, neurons form at least three layers – input, one or more hidden layers, and output. For each connection that occurs in a given layer, there is a certain real number called the weight of the connection. For the hidden layers and the output layer, each neuron uses a specific activation function. Based on the input values of a given neuron, this function determines its activation, i.e. a certain real value, which is the output value. As an activation function for more complex problems, it is advised to use a nonlinear function, as they can help to compute less trivial problems with less amount of neurons. Networks in which the value is propagated from input to output only are known as feed-forward networks. 
Having the basis of a multi-layer feed-forward network we can describe the learning process as changing the connection weights using a backpropagation algorithm i.e. propagating the errors back to the preceding layers as shown on Figure~\ref{fig:backpropagation_diagram} with help of gradient descent. Currently, there are many optimizers for the gradient descent, and one of the most recent and also used in this research is more memory efficient and overall fast- Adam optimizer \cite{KingmaB14}.

When the learning process is done the ANN can be used for predicting output parameters based on the given, possibly not seen by the network, input values.

\subsection{Motivation and research goals}

 The main aim of the research was to investigate how well can the ANN contribute to solving nuclear reactor design problems. This would enable reactor designers to consider using ANNs that can provide fast results and solutions before performing more detailed analysis on reactor design using more resource-consuming computations such as core simulator runs. Then, such a network was tested for performance in terms of different architectures and other parameters so the accuracy of its predictions is optimized and the network learning process is also optimal. In order for an artificial neural network to be taught the appropriate weights of neural connections, it is necessary to use the appropriate training data, which in the case of this research was generated by a Purdue Advanced Reactor Core Simulator (PARCS)~\cite{PARCS}.

\section{Methodology}
\label{sec:methods}

\subsection{General algorithm/procedure}
\begin{figure}[htp]
  \centering
  \includegraphics[width=1\textwidth]{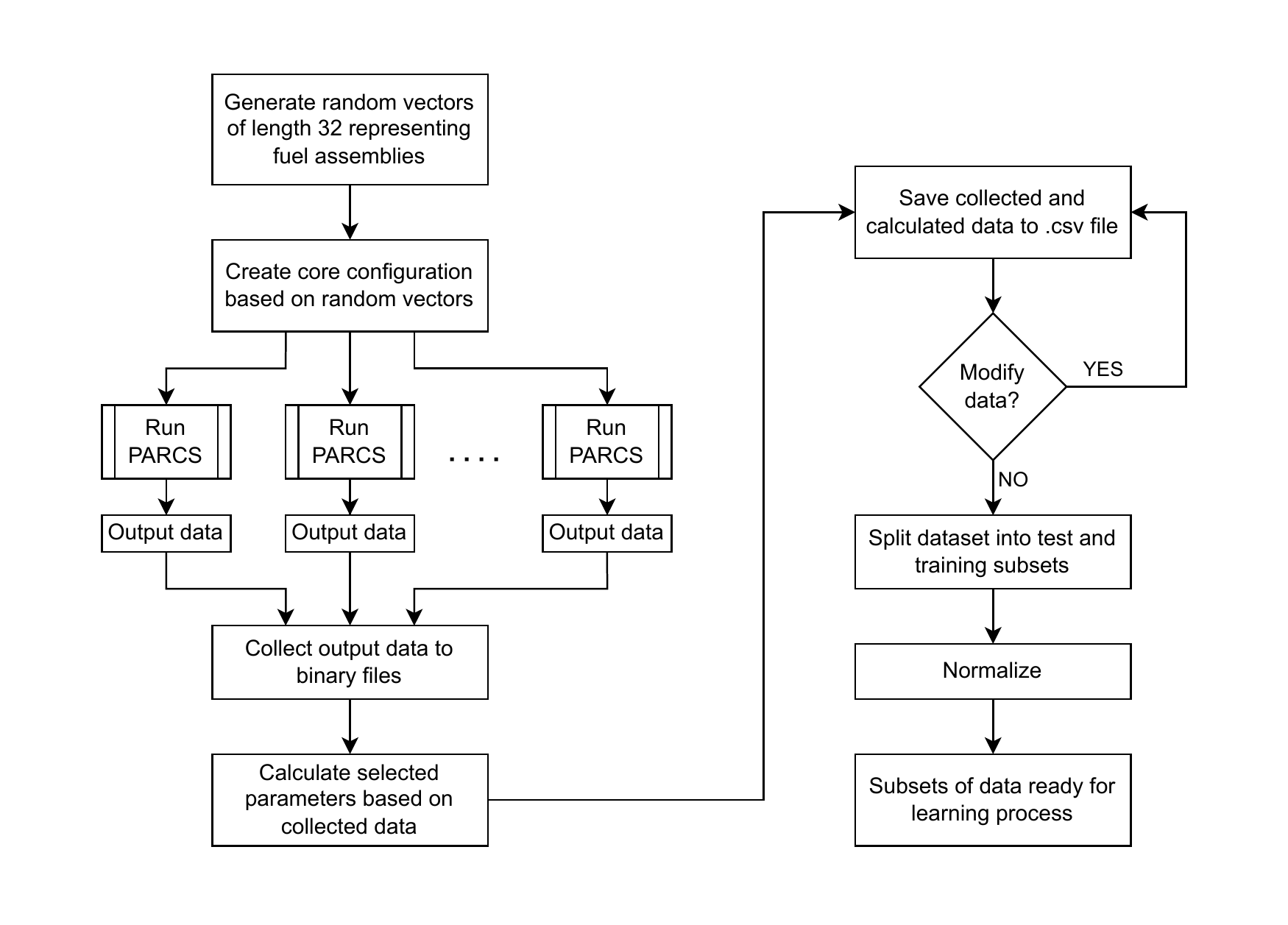}
  \caption{General workflow of the developed computational framework used for preparing data for ANN learning process including running the simulations, calculation of selected parameters, modification and transformation of data.}
  \label{fig:workflow_diagram_eng_7_5_in}
\end{figure}

For an easier understanding of the data generation procedure general workflow has been created and presented in the Figure~\ref{fig:workflow_diagram_eng_7_5_in}. A more detailed description of each step can be found in the next sections. 

\subsection{AI Technologies}
Due to the growing popularity of the Python programming language in machine learning and broadly understood data science, its ease of syntax, and a vast range of available libraries, it has been used as the primary programming language to create most of the code. Implementation of ANN was done using TensorFlow 2.0~\cite{tensorflow2015-whitepaper}. The additional power offered by the GPU (Graphics Processing Unit) was used to increase computing power when teaching the neural network through the CUDA platform supported by NVIDIA cuDNN (NVIDIA CUDA$^{\circledR}$ Deep Neural Network) \cite{CUDNN}. Using this method, the processor divides computing resources along with the graphics unit.

\subsection{Core Simulations and PARCS code}

In this work, a nuclear reactor core model \cite{darnowski2019analysis} was simulated with PARCS v3.2 core neutronics simulator \cite{PARCS}. It is a computer code being developed by the University of Michigan for U.S. Nuclear Regulator Commission \cite{parcs2}\cite{downar2012user}. PARCS is a popular tool used for nuclear reactor safety research by universities and governmental agencies. It simulates both steady-state type and transient type problems. It allows assessment of basic core parameters for the static case, e.g effective neutron multiplication factor, control rod worth, reactivity coefficients and other. It can be used to simulate slow and long term core changes like quasi-static fuel cycle or xenon transients. Moreover, it allows the simulation of more rapid transients with neutron kinetics phenomena like control rod ejection. Finally, it simulates neutron kinetics in a coupled mode with thermal-hydraulics system codes (like TRACE) during accidents (e.g. Main Steam Line Break or Anticipated Transient Without SCRAM). PARCS uses the nodal diffusion approach and dedicated numerical methods to relatively quickly solve neutron transport.

In this study, PARCS was used to find the effective neutron multiplication factor ($k_{eff}$) of a nuclear system as a function of long-term operation and fuel burnup (isotopic depletion) up to the end of a cycle. 
The search for $k_{eff}$ demands a solution of the static criticality problem, which is the eigenvalue problem for a reactor in a steady-state (or quasi steady-state) given by the general Equation~\ref{eq:static}, 

\begin{equation}\label{eq:static}
   M\phi = \frac{1}{k_{eff}} F\phi
\end{equation}

Where $\phi$ is neutron flux (eigenvector or eigenfunction), $k_{eff}$ is an eigenvalue, M is the migration matrix (operator) describing various neutron leaks and losses, F is the fission matrix (operator) responsible for the neutron generation. Forms of M and F matrices are dependent on the formulation of the problem and applied approximations (like diffusion approximation) used to solve the Neutron Transport Equation \cite{duderstadt1976nuclear}. 

For a nuclear system to be in a steady-state, a neutron population has to be time independent (constant). In the Equation~\ref{eq:static} losses given by the left hand side (LHS) have to be equal to the neutron production given by the right hand side (RHS). In a real physical nuclear system with a self-sustaining fission chain reaction (without external sources), the steady-state condition is only possible when RHS and LHS are equal to each other and $k_{eff}=1.0$. Otherwise, a system is in a transient state, supercritical when $k_{eff}>1.0$ and subcritical when  $k_{eff}<1.0$.

In the numerical analysis of nuclear systems, it is useful to solve static problem instead of the transient problem and calculate eigenvalue given by Equation~\ref{eq:static} with $k_{eff}\neq1.0$. In order to guarantee the correctness of Equation~\ref{eq:static}, the production term (RHS) has to be re-scaled by the scaling factor - eigenvalue ($k_{eff}$).
In effect, the eigenvalue quantifies the neutron multiplication potential of the system and practically it allows engineers to assess if a considered nuclear system is able to maintain a fission chain reaction.
To solve the problem PARCS uses the Wielandt eigenvalue shift method and Krylov CMFD solver \cite{parcs2}.

Furthermore, an essential neutronics-related parameter studied in this work is the so-called reactivity. It is defined as the net relative difference to the critical state ($\rho=0$ for $k_{eff}=1.0$) given by Equation~\ref{eq:rho}.

\begin{equation}\label{eq:rho}
    \rho = \frac{k_{eff}-1}{k_{eff}}
\end{equation}

 In this work, a fuel cycle length (measured in days) is considered as the time from the reactor's start-up with some initial excess reactivity level ($k_{eff}>1.0$) until the moment in time when excess reactivity of the core drops below zero ($k_{eff}\leq1.0$) due to fuel burnup and accumulation of neutron poisons. A self-sustaining nuclear fission chain reaction cannot be sustained in a system where the reactivity is below zero and from the practical point of view it means the end of a cycle. In practice, some parameters like power and temperature can be reduced to extend core operation time (stretched-out operation) but this situation is not studied in this work.

\subsection{BEAVRS Benchmark}
\begin{figure}[htp]
    \centering
    \includegraphics[width=0.7\textwidth]{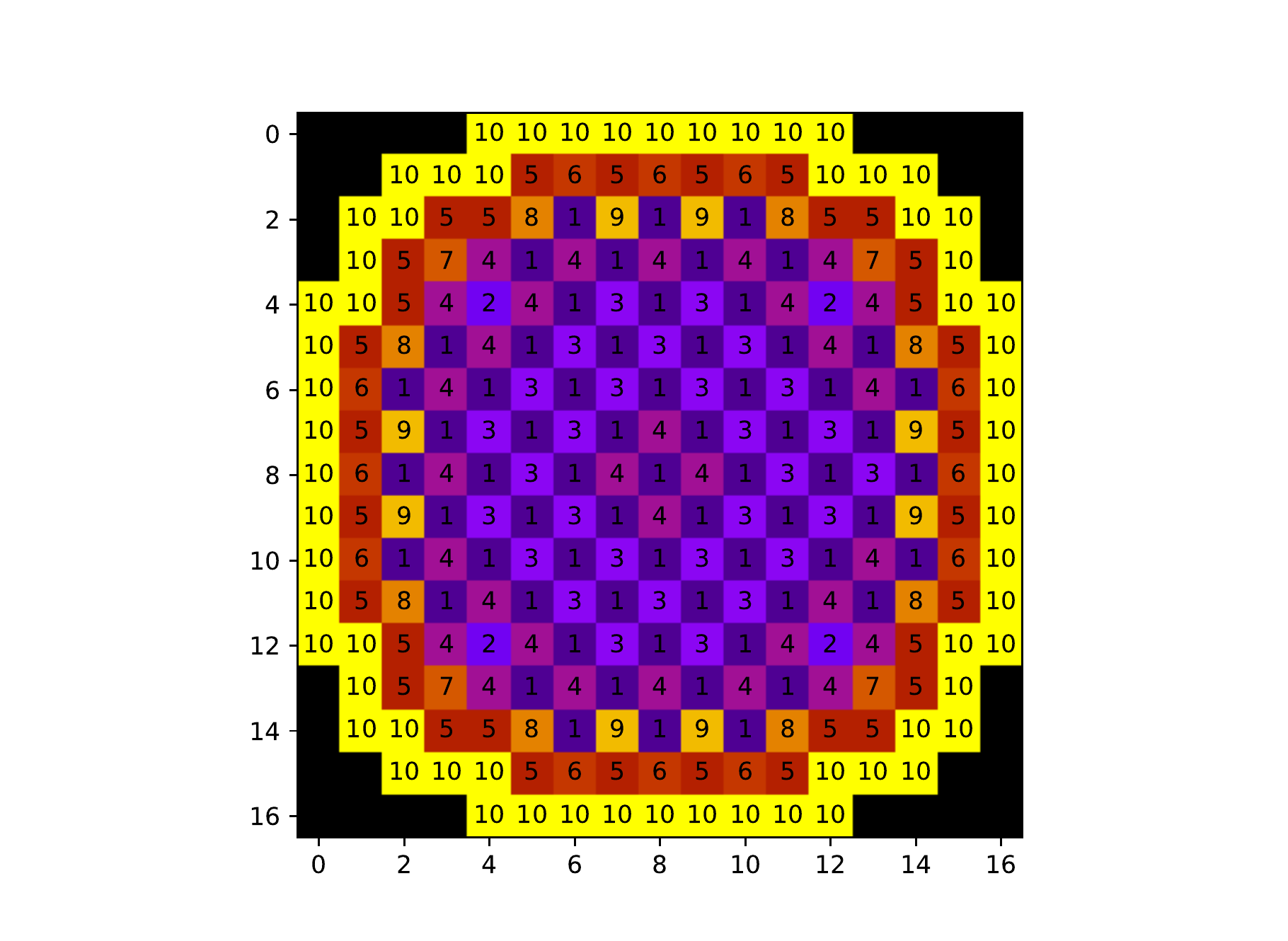}
    \caption{BEAVRS Benchmark reactor core. Based on \cite{darnowski2019analysis}.}
    \label{fig:beavrs_configuration}
\end{figure}

 The investigated  core is the Westinghouse 4-loop Pressurized Water Reactor (PWR) with thermal power equal to 3411~MWth. It was defined in the BEAVRS (Benchmark for Evaluation And Validation of Reactor Simulations) benchmark published by the MIT Computational Reactor Physics Group \cite{BEAVRS}.
 The BEAVRS first fuel cycle was studied, and its core loading pattern is presented in Figure \ref{fig:beavrs_configuration}. The first core contains nine fuel assembly types with 17x17 lattice design and three enrichments equal to 1.6, 2.4, 3.1 wt\%, different number of borosilicate Burnable Absorbers (BA) rods in assemblies: 0, 6, 12, 15, 16, 20. Details of fuel assemblies are presented in Table~\ref{tab:fuel_assemblies_description}.

\begin{table}[htp]
    \centering
     \caption{Details of the BEAVRS first fuel cycle fuel assemblies used in this work. }
    \begin{tabular}{cccc}
        \toprule
        Fuel assembly number & Name & Enrichment [\%] & Number of BA Rods    \\
        \midrule
        1 & FA1 &  1.6 &    0  \\
        2 & FA2 &  2.4 &    0  \\
        3 & FA3 &  2.4 &    12 \\
        4 & FA4 &  2.4 &    16 \\
        5 & FA5 &  3.1 &    0  \\
        6 & FA6 &  3.1 &    6  \\
        7 & FA7 &  3.1 &    15 \\
        8 & FA8 &  3.1 &    16 \\
        9 & FA9 &  3.1 &    20 \\
        10 & REFLECTOR &   &       \\
        \bottomrule \\
    \end{tabular}
   
    \label{tab:fuel_assemblies_description}
\end{table}

\subsection{Data generation}
\begin{figure}[htp]
  \centering
  \includegraphics[width=0.5\textwidth]{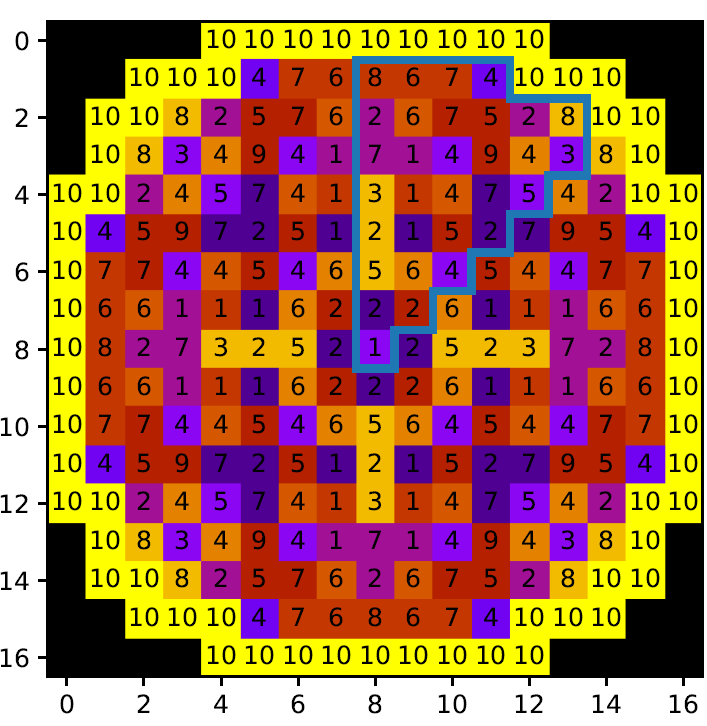}
  \caption{Example of generated configuration of the reactor core (with highlighted 1/8 symmetry).}
  \label{fig:sample_configurations}
\end{figure}

A fuel assembly configuration in a core is called a loading pattern. The selection of a loading pattern is an element in predicting its behavior in simulations using PARCS. In this work, it was assumed that the reactor core has the symmetry of $1/8th$. In the case of the tested core with a 17x17 assembly arrangement, a vector of 32 elements with a range of 1-9 was drawn. Ultimately, the number of different core configurations generated was equal to $N=10000$, of which an exemplary configuration is plotted in Figure~\ref{fig:sample_configurations}.

In order to accelerate data generation, PARCS was launched multi-process, which allowed for parallel simulations for many different configurations. It was achieved using a multiprocessing library for Python.

\subsection{Data gathering}
The data gathering process began with reading selected lines from PARCS output files. The read vectors containing the selected values were transposed to be placed as columns in the target matrix with values ready for the process of neural network learning. 

In the data matrix, each row was prepared to contain information for one separate configuration and its features based on $k_{eff}$ values, calculated using PARCS. Thus, the first 32 integers corresponding to the fuel assemblies represented the configuration of the randomly generated 1/8th of the core, following values described the evolution of $k_{eff}$. The last column contained the length of the cycle calculated as a linear interpolation between two burn-up steps for which the value of $k_{eff}$ drops below 1.0 (or $\rho<0$). This method seems to be justified as the behavior of the $k_{eff}$ is close to linear at the end of the cycle.           

Before starting the learning process, the input data was modified by conversion of numbers of fuel assemblies (integers 1-9) into cycle lengths with values corresponding to specific fuel assembly. In this way, the value "$1$" was changed to the value of cycle length for the core consisting only of assemblies of type "$1$", value "$2$" was changed to cycle length for core made only of assemblies "$2$" and so on. It was believed that it will improve the learning process, as after this conversion the input data could reflect, to some extent, features of the fuel assemblies used in the core. Also, as $k_{eff}$ usually does not deviate too much from 1.0, for easier assessment of the performance of the ANN and comparison of the results (especially relative values), reactivity (Equation: \ref{eq:rho}) instead of $k_{eff}$ is used in the rest of the study.

\subsection{Exploratory Data Analysis and data preparation}
\begin{figure}[htp]
  \centering
  \includegraphics[width=0.6\textwidth]{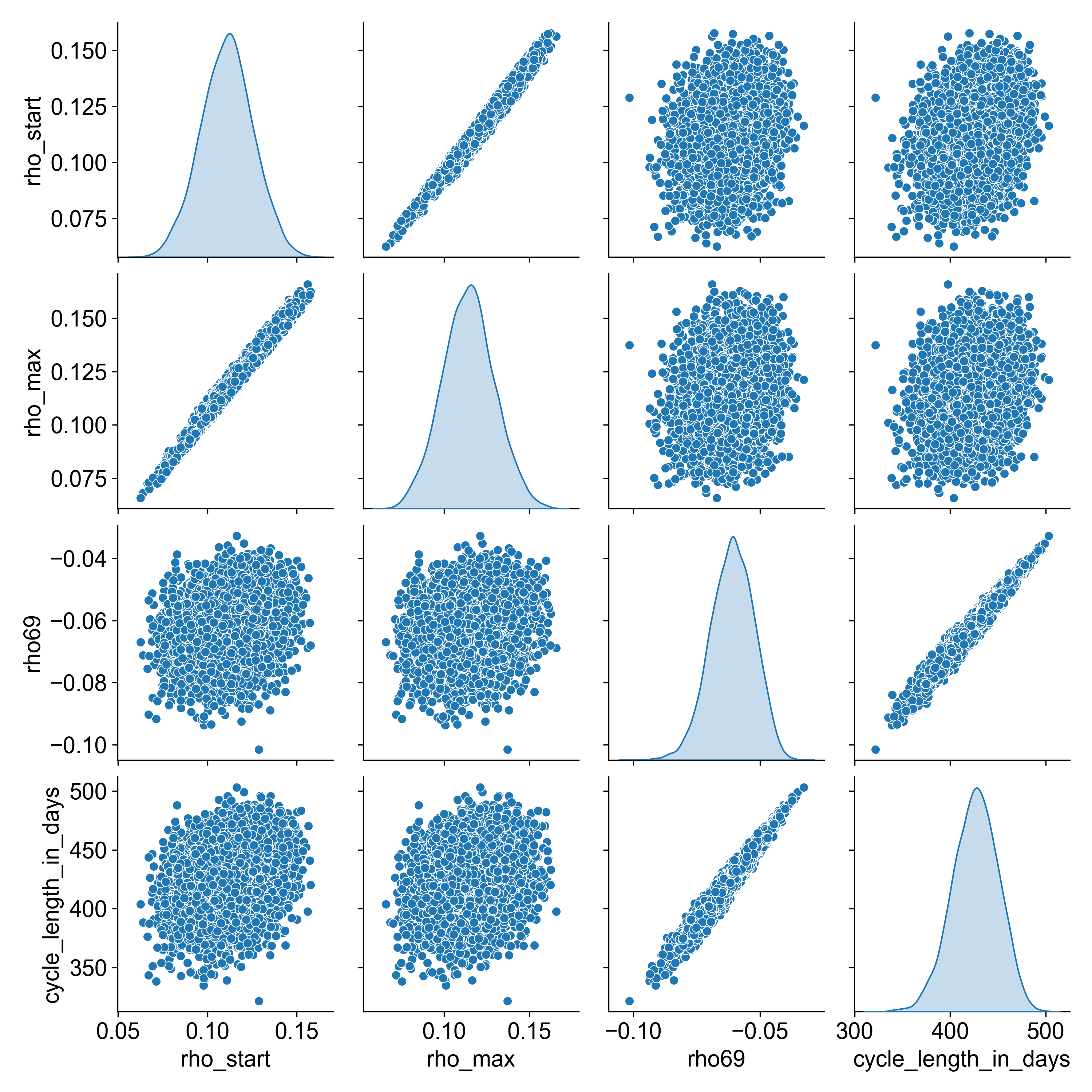}
  \caption{Pairplot of the data used for learning the ANN. Labels correspond to reactivity at the start of the cycle ($rho\_start$), maximal reactivity ($rho\_max$), reactivity close to the end of the cycle ($rho\_69$), and cycle length in days ($cycle\_length\_in\_days$).}
  \label{fig:data_pairplot}
\end{figure}

In order to get acquainted with the data set, some methods known from EDA (Exploratory Data Analysis) were used, such as the calculation of basic statistics and the visualization of the relationship between different parameters shown in Figure~\ref{fig:data_pairplot}.

At first glance, a visible linear correlation between reactivity at the start of the cycle and maximal reactivity is observed. It can be explained by checking the occurrence of maximal reactivity in a cycle which always occurs on the second step of the simulation. This is likely due to variations in the level of xenon and poisons, which can be observed in later stages of research when predicting reactivity progression in time. A second almost linear correlation is visible between cycle length and the reactivity at the end of the cycle. Again, it finds its explanation with how reactivity progress in time (after reaching maximum reactivity it shows decreasing linear trend in time). 

Data prepared for learning artificial neural network was split into 80/20 proportions and separated into features $x$ (columns of data matrix which can be used for learning)  and labels $y$ (output values that can be predicted). It was decided that the output of the ANN are values corresponding to normalized values of maximum reactivity, reactivity on the course of the cycle (time-dependency) applicable as $rho1, rho2, rho3$ and after that, every second reactivity value $rho4, rho6, rho8...$ and also the cycle length of the nuclear reactor core. Based on the values, the normalization layer of the neural network has been adapted, which is used to transform the data $x$ to obtain a dataset with a mean value $\mu=0$ and a standard deviation $\sigma=1$. This standardization layer was implemented using Keras API Layers~\cite{keras}. Due to the need of reversing the process of label normalization, an additional mechanism has been designed using the StandardScaler class from the \emph{ sklearn.preprocessing} library~\cite{scikit-learn}.

\subsection{Study of ANN architecture}\label{sec:ann_architecture}
One of the essential elements of ANN is its architecture – the number of layers, the number of neurons in each of them, and the activation functions for each layer. The issue of choosing the number of hidden layers is more straightforward than the selection of the optimal number of neurons in each of them due to the simple classification of problems in terms of the number of layers. Having only the input and output layers allows for solving linear problems such as, for example, the classification of linearly separable objects. Adding a single hidden layer makes it possible for a neural network to approximate virtually any function that maps input values continuously to finite solution space. A neural network with two hidden layers allows for solving complex problems without significant limitations. Based on~\cite{heaton2008introduction} there is currently no theoretical basis for a neural network to have more than two hidden layers, since two are sufficient for virtually any problem, due to this it was decided that the studied ANN would have two hidden layers. 

Having a fixed number of layers in the architecture is necessary to decide how many neurons are in each layer in combination with what activation function plays the best role in predicting the operation parameters of the nuclear reactor core. In addition, between the two hidden layers and between the last hidden layer and the output layer, a so-called dropout layer was added, resetting the input value of a given neuron at a particular frequency during network learning. The process of randomly zeroing certain connections in a neural network allows for achieving better accuracy of network predictions, which is presented in \cite{DBLP:journals/corr/abs-1207-0580}. 

The process of hyperparameter tuning was based on running the nested loops that ran the process of neural network learning with different combinations of neurons in different layers and dropout parameters. When studying the architecture of a neural network, it is also necessary to mention several processes, such as the initialization of connection weights or the selection of the appropriate activation function. For this reason, the initialization with the homogenous Glorot distribution, also called the Xavier distribution, presented in Equation \ref{eq:glorot_dist} was used ($n_{in}$ represents a number of neurons in the layer before, whereas $n_{out}$ represents a number of the following layer). The use of such a distribution means that the selected layers will not contribute too much to the learning process, as all layers will have a similar contribution \cite{pmlr-v9-glorot10a}.
\begin{equation}\label{eq:glorot_dist}
    W \sim Uniform\bigg(-\frac{\sqrt{6}}{\sqrt{n_{in}+n_{out}}}, \frac{\sqrt{6}}{\sqrt{n_{in}+n_{out}}}\bigg)
\end{equation}
As an activation function, the Gaussian Error Linear Units (GELU) was chosen for every layer, as in \cite{DBLP:journals/corr/HendrycksG16} it was suggested that GELU performs better than ReLU and ELU activation functions. 

\section{Results and discussion}
\label{sec:results}

\subsection{Analysis of the data gathering process}
The process of data generation resulted in 10 000 random pattern calculations with about 70 GB of data (PARCS simulations output data). The data collection process contributed to the creation of a 9.82 MB dataset with modified fuel assembly information (fuel cycle length for core consisting of only one corresponding assembly type). The memory occupied by both datasets was only approximately 0.04\% of the simulation output files, representing a considerable reduction in the size of the data that would have to be read from many different directories to create training data for a neural network. It should be taken into account that the data collection process consisted of extracting only part of the information contained in the simulation output files, so it is more about organizing and structuring the training data rather than compression.

\subsection{Optimal ANN architecture analysis}
\begin{figure}[htp]
    \centering
    \begin{subfigure}{0.5\textwidth}
        \centering
        \includegraphics[width=0.99\linewidth]{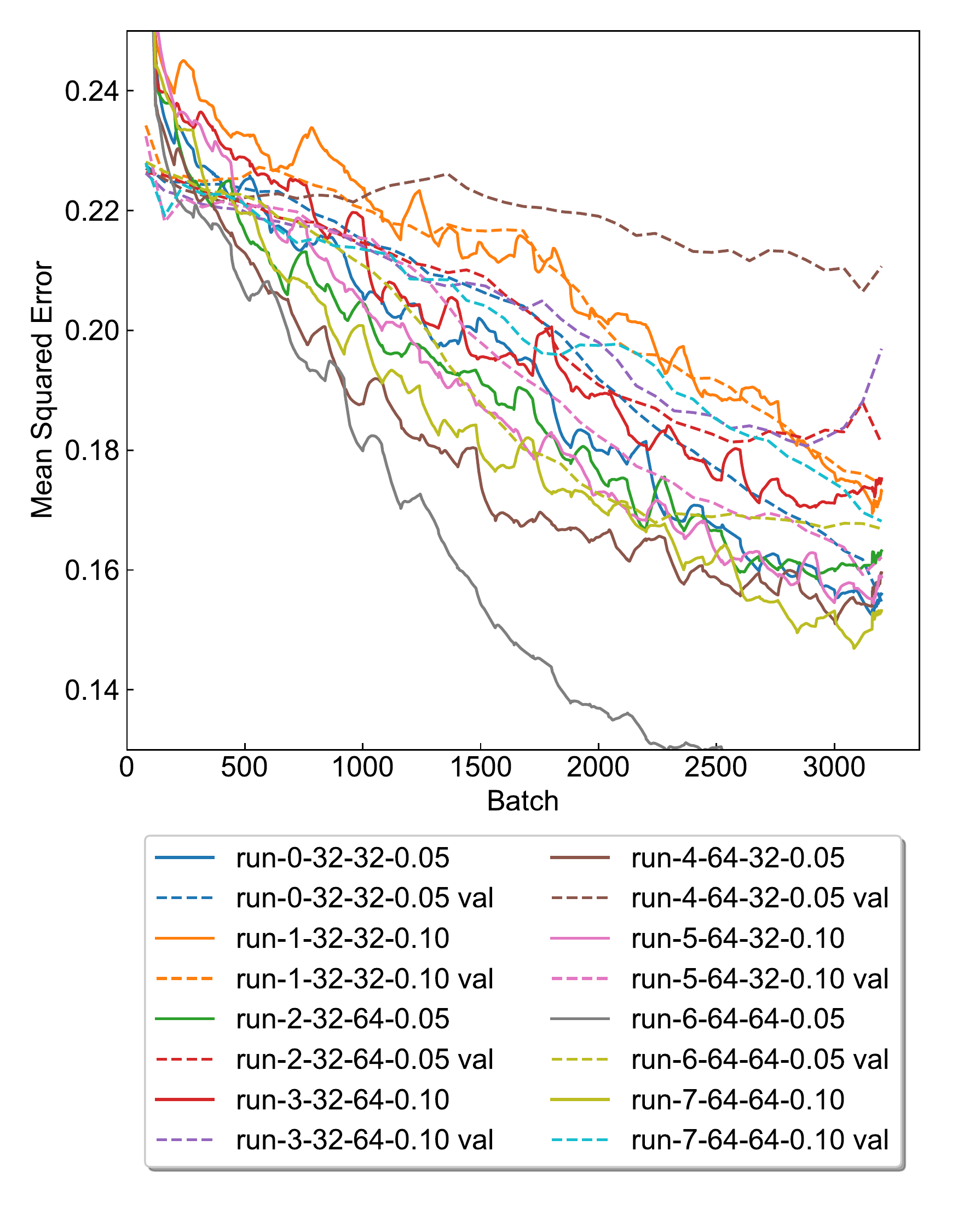}
        \caption{32, 64 neurons combination}
        \label{fig:MSE_vs_batch_32_vs_64}
    \end{subfigure}%
    \begin{subfigure}{0.5\textwidth}
        \centering
        \includegraphics[width=0.99\linewidth]{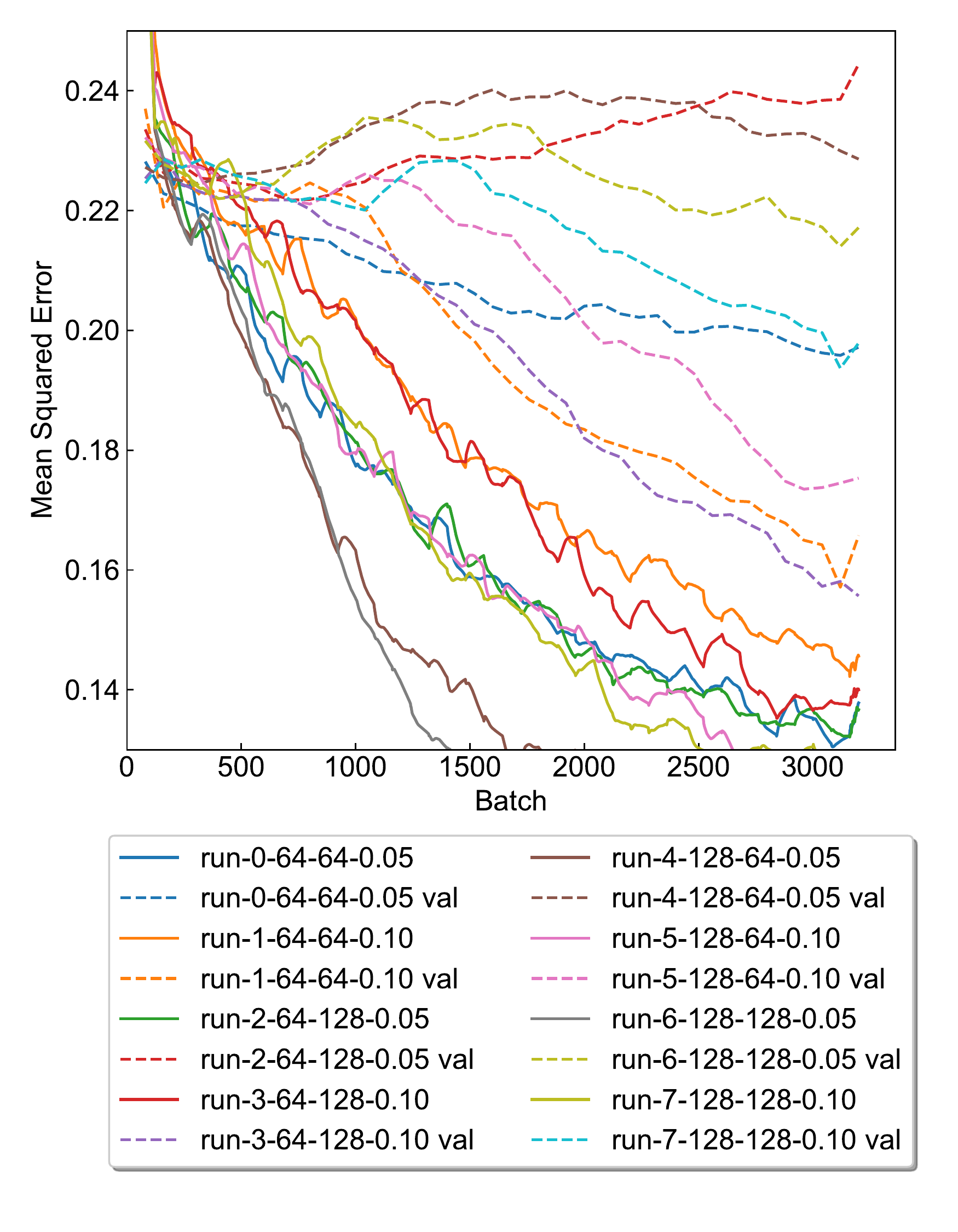}
        \caption{64, 128 neurons combination}
        \label{fig:MSE_vs_batch_64_vs_128}
    \end{subfigure}
\caption{Process of learning the ANN shown as Mean Squared Error (MSE) in the function of a batch number.}
\label{fig:MSE_vs_batch}
\end{figure}

Based on the ANN learning process for different hyperparameters, an attempt was made to determine which configuration works best for predicting the operation parameters of the considered PWR reactor core. For this purpose, the dependence of the value of the loss function on subsequent batches was prepared, shown in Figure~\ref{fig:MSE_vs_batch_32_vs_64} and Figure~\ref{fig:MSE_vs_batch_64_vs_128}, where each the launch of a given combination has a name with a given format:
\begin{equation}
    run-\alpha-NH1-NH2-\delta \; (val),
\end{equation} 
with optional note "val" indicating whether the loss function is calculated for validation (test) data or for teaching data in a given epoch, where $\alpha$ is a number denoting the running combination of hyperparameters, $NH1$ is equal to the number of neurons in the first hidden layer, $NH2$ is equal to the number of neurons in the second hidden layer and $\delta$ indicates the frequency of dropout (abandonment of a given connection) defined as a percentage of connections which weights will be set to zero in each update cycle.

\begin{figure}[htp]
    \centering
    \includegraphics[width=0.8\textwidth]{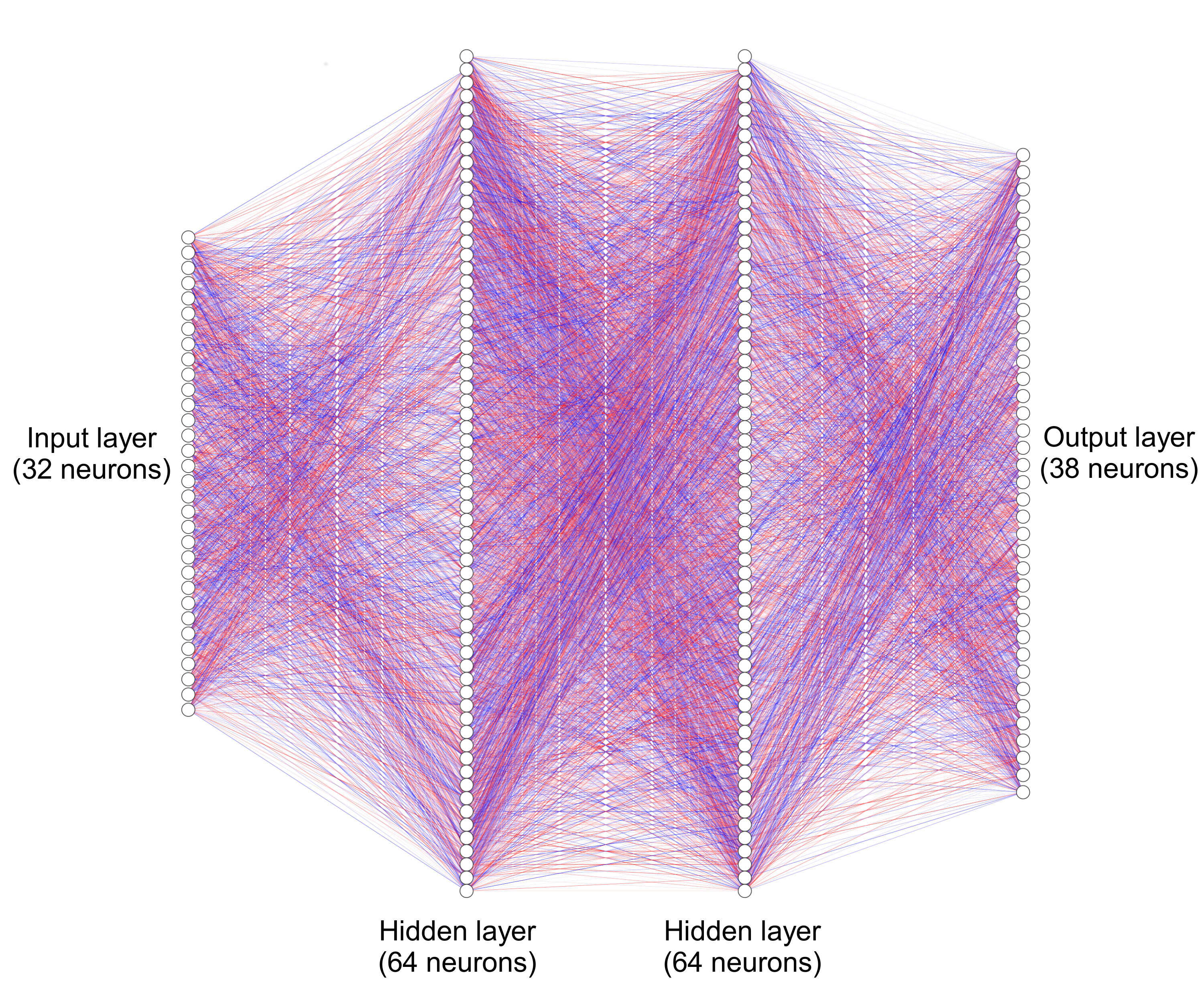}
    \caption{Visualization of the studied artificial neural network. Created using NN-SVG visualizer \cite{LeNail2019}.}
    \label{fig:NN-64-64}
\end{figure}

Based on Figure~\ref{fig:MSE_vs_batch_32_vs_64} and Figure~\ref{fig:MSE_vs_batch_64_vs_128}, it is difficult to determine the best architecture of the ANN. This is largely due to certain randomness of the learning process resulting from mixing the dataset and the chances of resetting important connections through dropout. For some architectures, one can notice an early problem of the so-called overfitting - a problem for which the weights of connections are fitted too much to match the learning dataset. As a result, the test data is recognized incorrectly (with lower accuracy). An example of such an architecture is a network consisting of 64 neurons in the first hidden layer and 128 neurons in the second hidden layer with a dropout value of 0.05 (run-64-128-0.05). Again, it is necessary to mention certain randomness in network learning. The loss function of this architecture can also be so high due to the unfortunate dropout of important connections resulting from the equal probability of choosing each connection. 
Due to the thriving architecture consisting of 64 neurons in both hidden layers (one of the lower values of the loss function and the slight difference between the value of the loss function for the learning set and the test set), it was decided to choose this architecture to analyze the prediction of the parameters of the nuclear reactor core at a dropout value of 0.1. The selected neural network is shown in Figure~\ref{fig:NN-64-64}, where the colors of the connections denote the initialized initial value, and the additional neuron in two hidden layers represents bias.

\subsection{Prediction of selected core parameters}
\begin{figure}[htp]
    \centering
    \includegraphics[width=0.7\textwidth]{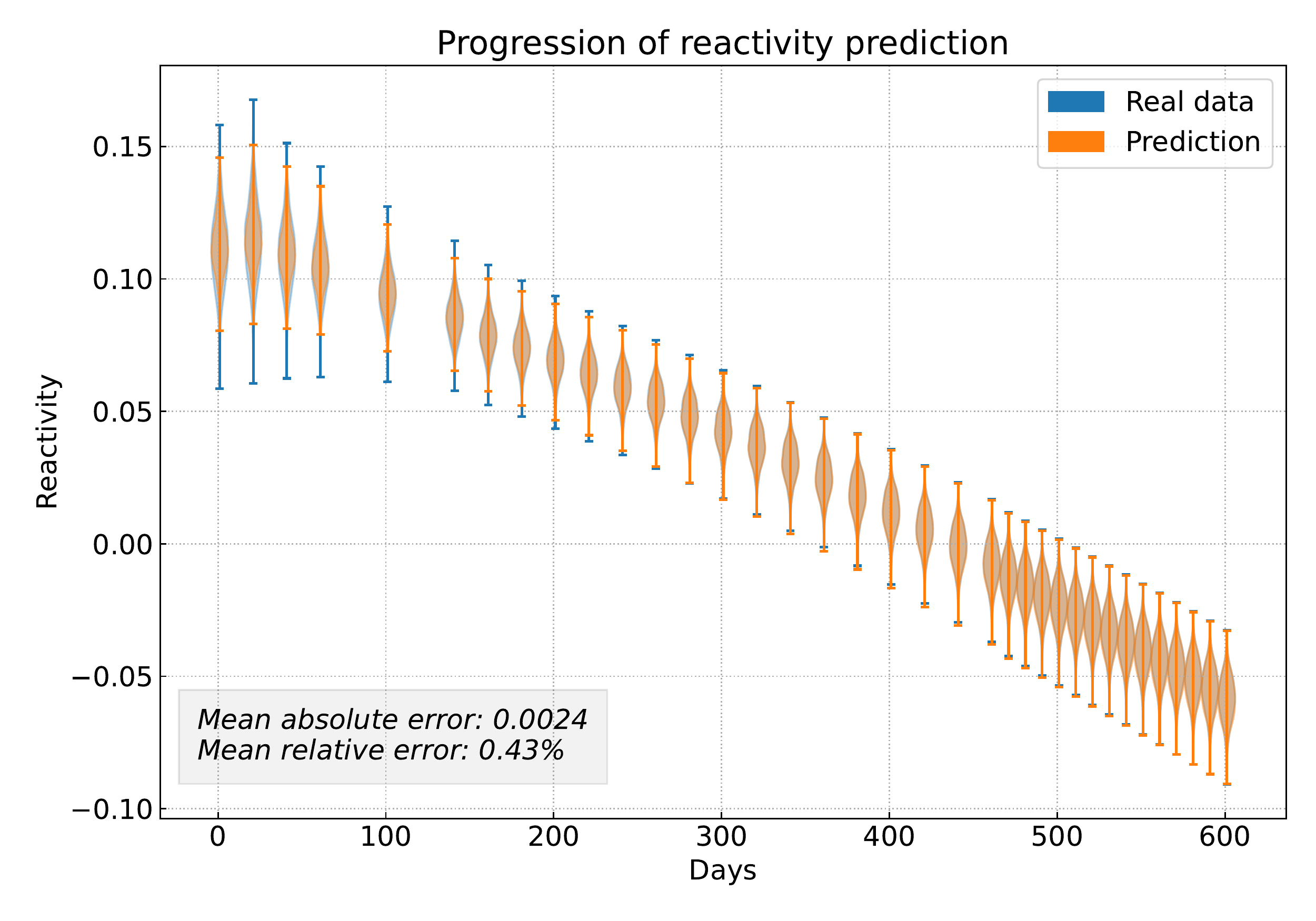}
    \caption{Dependence on the day of the fuel cycle of core reactivity for PARCS calculated (real) and ANN (predicted) calculated data.}
    \label{fig:reactivity_vs_days}
\end{figure}

Having the ANN with 32 neurons in the input layer, 64 in each hidden layer, and 38 neurons in the output layer, it was possible to scale normalized values of reactivity back to their original values and obtain the reactivity dependence of time in the fuel cycle shown on Figure \ref{fig:reactivity_vs_days}.
The average relative error of the prediction was 0.43\%, and the absolute error was 0.0024. 
It is noticeable that on the initial days of the cycle, the deviation is different for real and predicted data compared to other days of the cycle, where the prediction coincides very well with the actual data. This is likely due to variations in the level of xenon and poisons, which can be very nonlinear and thus hard to predict. 
It can suggest putting more emphasis on the first half of the cycle in future works. As the processes at the beginning of the cycle are more complex they may require denser sampling or an additional, more advanced model. 

\begin{figure}[htp]
    \centering
    \includegraphics[height=0.35\textheight]{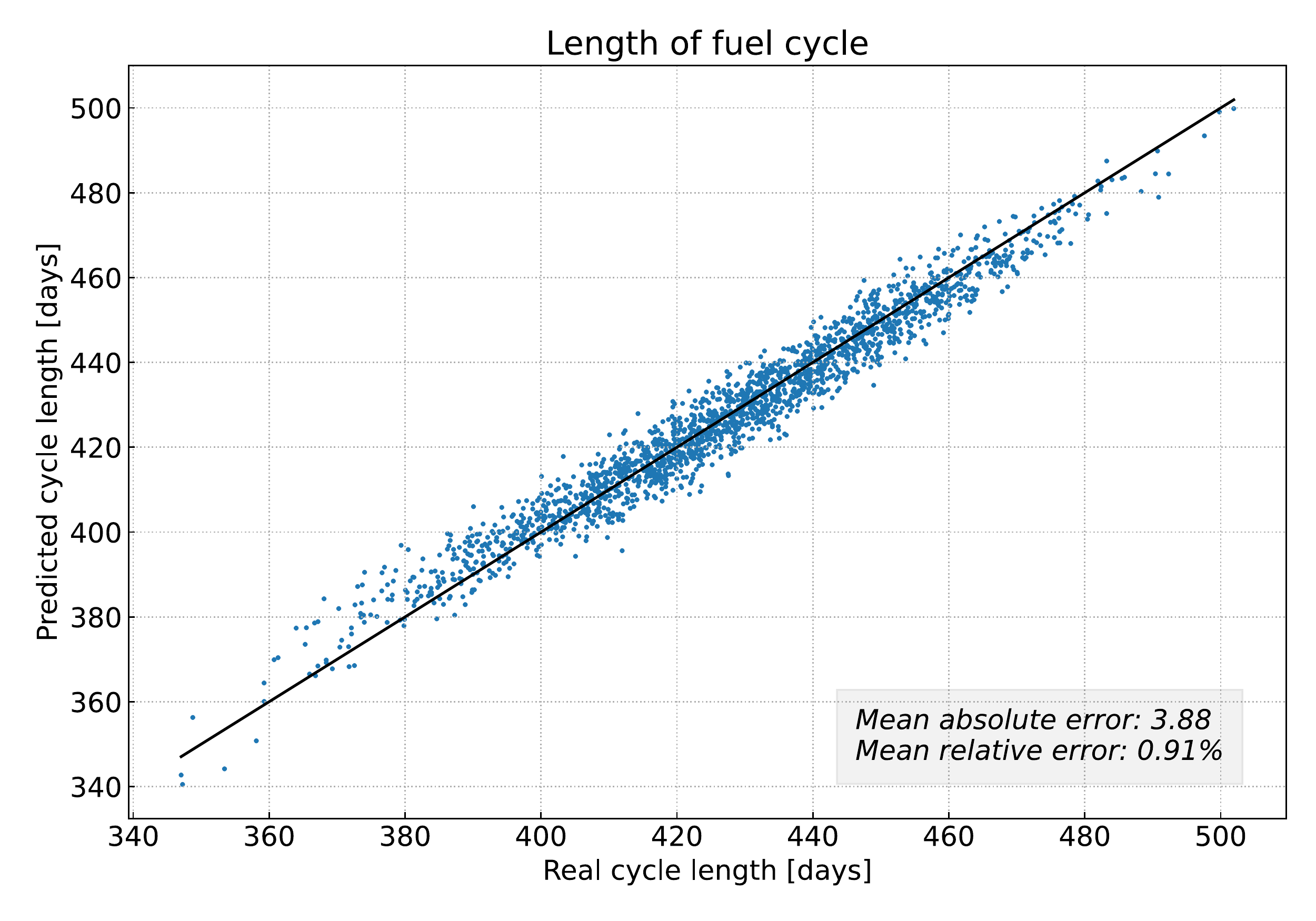}
    \caption{Predicted fuel cycle length compared to the real value.}
    \label{fig:cycle_length}
\end{figure}%

\noindent Errors for the collation of fuel cycle length actual and predicted values is characterized by a level of 0.91\% as for relative error and 3.88 days for an absolute error, which indicates a very good accuracy of predictions. In addition, for this data, the Pearson correlation coefficient was calculated at $\rho\approx 0.98$, which confirms a relationship of predicted to actual values close to linear. 
Comparing the uncertainties resulting from the use of a simulator taking into account neutron diffusion and the uncertainties of nuclear data, it can be said that the data obtained based on neural network predictions have very similar or sometimes lower levels of uncertainty values. 

\begin{figure}[htp]
    \centering
    \includegraphics[width=0.75\textwidth]{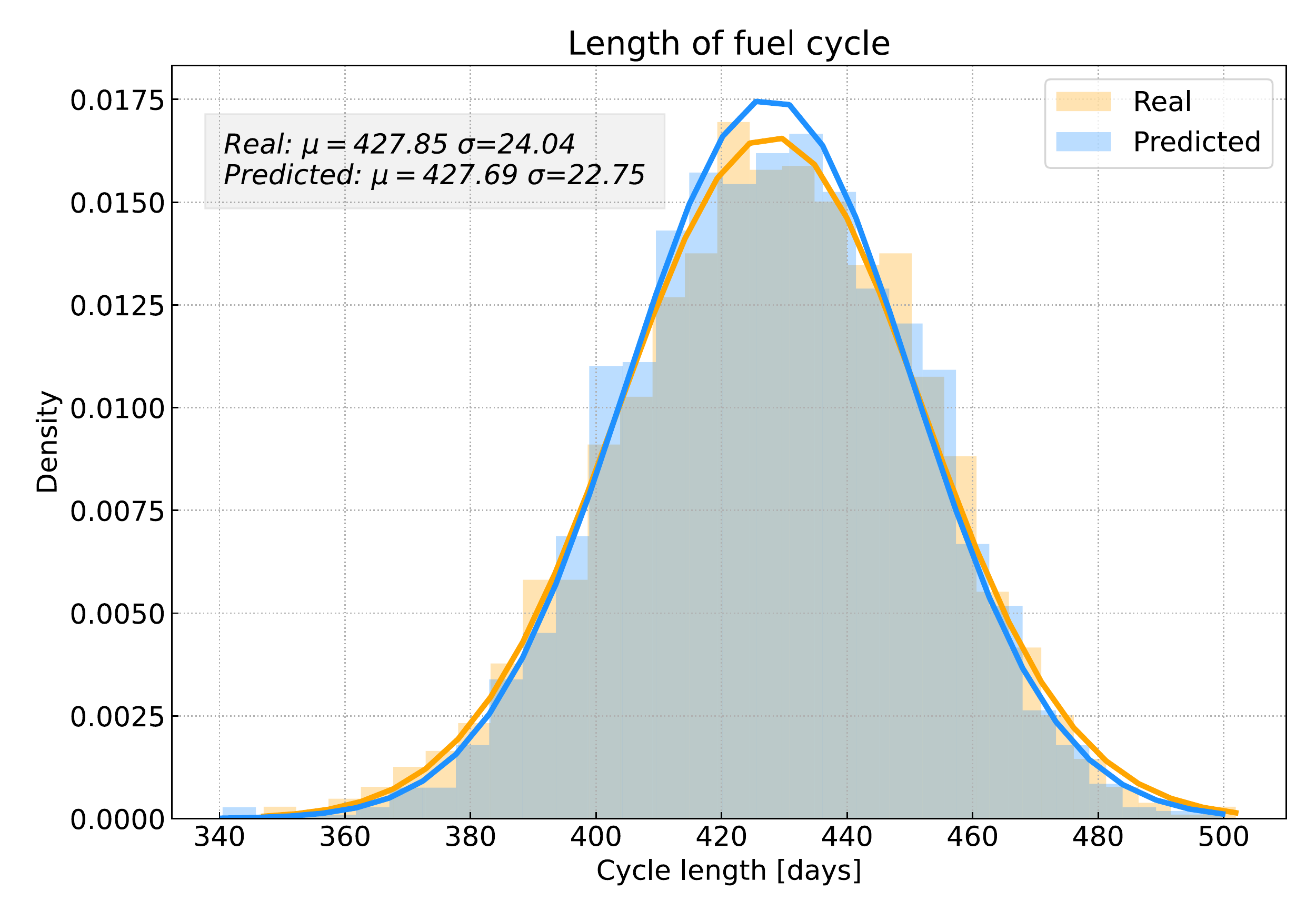}
    \caption{Distribution of the real and predicted values of fuel cycle length.}
    \label{fig:cycle_length_histogram}
\end{figure}

In addition, in order to test the accuracy of the predicted values of the core duty cycle, a histogram was prepared to compare the distribution of actual and predicted values shown in Figure~\ref{fig:cycle_length_histogram}. Normal distribution curves were matched to histograms, from which information about the mean $\mu$ and standard deviation $\sigma$ was obtained. 

The analysis of the fuel cycle length distributions shows that these distributions have very similar parameters. The average value is approximately equal, and the deviation value stands out only slightly. The similarity of the distributions indicates, to some extent, a good prediction of the ANN. 

\section{Conclusions and Summary}

The aim of this work was to create a design and implementation of an artificial neural network that can calculate selected operating parameters of the nuclear reactor core. The study of the network architecture resulted in similar results for many different architectures, which did not allow unambiguously determining the best number of neurons in each layer. In the case of some combinations, the phenomenon of so-called overfitting, i.e., too much matching with the learning data, resulting in a worse result for the loss function for test data, was noticed. Finally, it was decided that the best neural network architecture for the problem of predicting the operating parameters of the nuclear reactor core is a neural network with 64 neurons in each hidden layer, and the dropout value is set at 0.1 between the hidden layers and the hidden layer and the output layer. The research provides insight into the possibility of using artificial neural networks for issues related to nuclear energy. It leaves a wide range of further research, such as the improvement of the learning dataset, which, with the right amount of information, can improve the accuracy of predictions and the possibility of adding other predicted parameters such as the concentration of poisons in the nuclear reactor core or the temperature of the fuel. This process could allow the creation of a core simulator based on an artificial neural network, support the design of core configurations in terms of proper arrangement of fuel assemblies, or optimization of the fuel cycle under certain conditions. What is more, the database created from the PARCS output files has a huge amount of parameters that were not used in these studies, which in the long run may allow them to be used to increase the accuracy of predictions as well as the ability to calculate more nuclear reactor core parameters.

\section{Acknowledgments}
Research was funded by Warsaw University of Technology within the Excellence Initiative: Research University (IDUB) programme.

Source code used for the research may be found at
  \url{https://github.com/dazeeeed/neural-physics}
under the GNU GPL v3.0 license.

\bibliographystyle{unsrt}  
\bibliography{references}

\end{document}